\documentclass{article}

\PassOptionsToPackage{numbers, compress}{natbib}

\usepackage[preprint]{neurips_2024}

\usepackage[utf8]{inputenc} 
\usepackage[T1]{fontenc}    
\usepackage{hyperref}       
\usepackage{url}            
\usepackage{booktabs}       
\usepackage{amsfonts}       
\usepackage{nicefrac}       
\usepackage{microtype}      
\usepackage{xcolor}         
\usepackage[most]{tcolorbox}
\usepackage{amssymb}
\usepackage{wrapfig}
\usepackage{enumitem}
\usepackage{siunitx}
\usepackage[acronym]{glossaries}
\usepackage{etoc} 
\makeglossaries

\glsdisablehyper
\newacronym{soohak}{SH$^2$}{SH$^2$ mathematics benchmark}

\let\svthefootnote\thefootnote
\newcommand\freefootnote[1]{%
  \let\thefootnote\relax%
  \footnotetext{#1}%
  \let\thefootnote\svthefootnote%
}

\usepackage{kotex}

\definecolor{citecolor}{HTML}{0071bc}
\definecolor{cadmiumgreen}{rgb}{0.0, 0.42, 0.24}
\definecolor{cornellred}{rgb}{0.7, 0.11, 0.11}
\hypersetup{
  colorlinks = true,
  linkcolor  = cornellred,
  citecolor  = citecolor,
  urlcolor   = cadmiumgreen,
}

\newcommand{\soohakmini}{SOOHAK-Mini}
\newcommand{\soohakfull}{SOOHAK}
\newcommand{\shmini}{\soohakmini}
\newcommand{\shfull}{\soohakfull}
\newcommand{\challengesplit}{Challenge split}

\title{\textsc{Soohak}: A Mathematician-Curated Benchmark for Evaluating Research-level Math Capabilities of LLMs} 

\begin{document}

\maketitle

\vspace{-2cm}

\subsection*{Organizing Team}
{\fontsize{8.5}{8}\selectfont
Guijin Son$^{1,2,21}$, 
Seungone Kim$^{3}$,
Catherine Arnett$^{2}$,   
Hyunwoo Ko$^{1}$, 
Hyein Lee$^{5}$,
Hyeonah Kang$^{34}$,
Jiang Longxi$^{33}$,
Jin Yun$^{33}$,
JungYup Lee$^{4}$,
Kyungmin Lee$^{5}$,
Sam Yoosuk Kim$^{6}$,
Sang Park$^{4}$,
Seunghyeok Hong$^{30}$,
SeungJae Lee$^{4}$,
Seungyeop Yi$^{21}$,
Shinae Shin$^{34}$, 
SunHye Bok$^{6}$,
Sunyoung Shin$^{34}$,
Yonghoon Ji$^{6}$,
Youngtaek Kim$^{5}$,
Hanearl Jung$^{1}$,
Akari Asai$^{3}$,
Graham Neubig$^{3}$,
Sean Welleck$^{3}$,
Youngjae Yu$^{21}$
}

\par\vspace{0.2em}
{\fontsize{8.5}{8}\selectfont
For questions or model-evaluation requests, contact \href{mailto:guijin.son@snu.ac.kr}{\texttt{guijin.son@snu.ac.kr}}.
\par}


\newcounter{daggerfootnote}
\newcommand*{\daggerfootnote}[1]{%
    \setcounter{daggerfootnote}{\value{footnote}}%
    \renewcommand*{\thefootnote}{\fnsymbol{footnote}}%
    \footnote[2]{#1}%
    \setcounter{footnote}{\value{daggerfootnote}}%
    \renewcommand*{\thefootnote}{\arabic{footnote}}%
}

\renewcommand{\thefootnote}{\fnsymbol{footnote}}
\subsection*{Dataset Contributors\daggerfootnote{Names in alphabetical order by given name.}}


{\fontsize{8.5}{8}\selectfont 
Akshelin R$^{7}$, 
Alexander B. Ivanov$^{8}$, 
Boboev Muhammadjon$^{9}$, 
Chae Young Han$^{10}$,
Christian Stump$^{8}$,
Cooper R. Anderson$^{31}$,
Dmitrii Karp$^{11}$,
Dohyun Kwon$^{12}$,
Dongryung Yi$^{12}$,
DoYong Kwon$^{13}$,
Duk-Soon Oh$^{14}$,
Eunho Choi$^{21}$,
Giovanni Resta$^{15}$,
Greta Panova$^{16}$,
Huiyun Noh$^{12}$,
Hyungryul Baik$^{12}$,
Hyungsun Bae$^{1}$,
Inomov Mashrafdzhon$^{17}$,
Jeewon Kim$^{12}$,
Jeong-Rae Kim$^{35}$,
Ji Eun Lee$^{18}$,
Jiaqi Liu$^{19}$,
Jieui Kang$^{20}$,
Jimin Kim$^{21}$,
Jon-Lark Kim$^{22}$,
Joonyeong Won$^{20}$,
Junseo Yoon$^{21}$,
Junwoo Jo$^{12}$,
Kibeom Kim$^{12}$,
Kiwoon Kwon$^{23}$,
Mario Kummer$^{24}$,
Max Mercer$^{25}$,
Min Hoon Kim$^{20}$,
Minjun Kim$^{21}$,
Nahyun Lee$^{26}$,
Ng Ze-An$^{27}$,
Nicolas Libedinsky$^{36}$,
Rafał Marcin Łochowski$^{28}$,
Raphaël Lachièze-Rey$^{29}$,
Robert Auffarth$^{36}$,
Ruichen Zhang$^{19}$,
Sejin Park$^{21}$,
Seonguk Seo$^{21}$,
Shin Jaehoon$^{21}$,
Sunatullo$^{31}$,
Taewoong Eom$^{21}$,
Yeachan Park$^{18}$,
Yongseok Jang$^{13}$,
Youchan Oh$^{21}$,
Zhaoyang Wang$^{19}$,
Zoltán Kovács$^{32}$
}



\vspace{1em}

\renewcommand{\thefootnote}{\arabic{footnote}}

\begin{abstract}

Following the recent achievement of gold-medal performance on the IMO by frontier LLMs, the community is searching for the next meaningful and challenging target for measuring LLM reasoning. Whereas olympiad-style problems measure step-by-step reasoning alone, research-level problems use such reasoning to advance the frontier of mathematical knowledge itself, emerging as a compelling alternative. Yet research-level math benchmarks remain scarce because such problems are difficult to source (\textit{e.g.}, Riemann Bench and FrontierMath-Tier 4 contain 25 and 50 problems, respectively). To support reliable evaluation of next-generation frontier models, we introduce \textsc{Soohak}, a 439-problem benchmark newly authored from scratch by 64 mathematicians. \textsc{Soohak} comprises two subsets. On the Challenge subset, frontier models including Gemini-3-Pro, GPT-5, and Claude-Opus-4.5 reach 30.4\%, 26.4\%, and 10.4\% respectively, leaving substantial headroom, while leading open-weight models such as Qwen3-235B, GPT-OSS-120B, and Kimi-2.5 remain below 15\%. Notably, beyond standard problem solving, \textsc{Soohak} introduces a refusal subset that probes a capability intrinsic to research mathematics: recognizing ill-posed problems and pausing rather than producing confident but unjustified answers. On this subset, no model exceeds 50\%, identifying refusal as a new optimization target that current models do not directly address. To prevent contamination, the dataset will be publicly released in late 2026, with model evaluations available upon request in the interim.

\end{abstract}

\section{Introduction}

Mathematical reasoning benchmarks offer a sharp probe of large language model (LLM) abilities, because they stress multi-step inference and precise final answers, spanning tasks from contest-style problem solving~\citep{aime, an2025amo} to research-adjacent questions~\citep{glazer2024frontiermath, schmitt2025improofbench}. Scraping questions from publicly available competitions and textbooks remains the dominant assembly route~\citep{gao2025omnimath}. It scales quickly, but increases training-data overlap and accelerates saturation as frontier models improve~\citep{balunovic_matharena_2025}. While human-authoring fresh problems sidesteps contamination, such efforts are typically confined to a single mathematical area (e.g., AMO-Bench~\citep{an2025amo}) or kept very small to remain tractable (e.g., Riemann-Bench~\citep{garre2026riemann}). This narrowness in field or size makes it difficult to compare models across difficulty levels or to localize where capability gaps lie. Moreover, the most recent generation of benchmarks responds to leakage by withholding problems behind access control~\citep{phan2025humanity, schmitt2025improofbench, ma2026eternalmath, abouzaid2026first}, which reduces contamination but trades away transparency and reproducibility. These pressures compound when evaluation must guide high-stakes pre-training and post-training initiatives, where benchmark integrity, breadth, and accountability all matter at the same time.

In this work, we introduce \shfull{}, consisting of a 340-item Challenge subset and a 99-item Refusal subset. The Challenge subset contains graduate level and research-adjacent material authored by 68 contributors, including 38 faculty members, 25 PhD students or postdoctoral researchers, and 5 master's or undergraduate IMO medalists. Across eleven closed and open-weight systems, \texttt{Gemini-3-Pro}, \texttt{GPT-5}, and \texttt{Claude-Opus-4.5} reach Avg@3 of 30.39\%, 26.37\%, and 10.39\% on \shfull{} Challenge. \texttt{Kimi-2.5} is the best open-weight model at 13.87\%. The Refusal subset evaluates whether models identify ill-posed prompts instead of producing confident answers. The best closed model reaches only 43.10\% Avg@3, while \texttt{GLM-5} reaches the highest score at 49.49\%, indicating that many systems continue attempting to solve prompts that are invalid as written. Additionally, to provide a subset for tracking smaller and open-weight systems, we release \shmini{}, a 702-question collection authored by a broader pool of 105 mathematicians and students. \shmini{} covers high-school olympiad through early graduate material. GPT-5 reaches the strongest \shmini{} Avg@3 at 72.22\%, while \texttt{Kimi-K2.5} is the strongest open-weight model at 66.07\%. The full collection is temporarily embargoed, with public release planned in late 2026\footnote{The dataset will be released before the NeurIPS 2026 final acceptance.}. In the interim, we evaluate models upon request.

Finally, we conduct a human baseline with 25 participants across five teams. The invited participants include IMO honorable mention recipients to gold-medalists, mathematically trained undergraduates, and PhD-level researchers in mathematics and computer science. On 79 prompts, aggregated teams cover 50.6\% of the sample, confirming that the benchmark is challenging but tractable for strong human solvers. We hope this baseline helps future work interpret model scores against skilled human coverage across expertise profiles.

\section{Related Work}





Benchmarks such as MATH~\citep{hendrycks2021measuring} and GSM8K~\citep{cobbe2021training} were among the earliest standardized evaluations of mathematical reasoning in LLMs. At the time of release, language models performed extremely poorly ($<$10\% accuracy). As model capabilities improve, these benchmarks have become less discriminative at the frontier, motivating a wave of newer math benchmarks designed to track rapid progress and, in some cases, to resist fast saturation~\citep{bytedance_seed_2025_beyondaime, phan2025humanity}.

The topic and difficulty of math benchmarks fall into two broad categories~\citep{an2025amo}. \emph{Olympiad-style} benchmarks emphasize multi-step problem solving in knowledge-contained settings, without requiring specialized background beyond standard contest curricula. They often admit short, machine-checkable final answers and are frequently derived from competition materials \citep{aime, HMMT2025, ko2025understand} or curated into unified suites~\citep{gao2025omnimath}. In contrast, \emph{research-level} benchmarks aim to probe advanced mathematical knowledge and longer-horizon reasoning, drawing on research literature or researcher-authored questions, as in FrontierMath~\citep{glazer2024frontiermath}, RealMath~\citep{zhang2025realmath}, and more recently First Proof~\citep{abouzaid2026first}.

Dataset construction choices also interact strongly with contamination risk. A large fraction of benchmarks are assembled from publicly available exams and competitions or from published sources~\citep{hendrycks2021measuring, cobbe2021training, gao2025omnimath, balunovic_matharena_2025, zhang2025realmath}. But items sourced from exams are vulnerable to overlap with training data, and contamination has been documented in widely used contest-derived sets~\citep{balunovic_matharena_2025}. Once contaminated, benchmark scores can substantially overestimate true generalization~\citep{glazer2024frontiermath}. To mitigate these issues, some efforts rely on newly written questions or carefully controlled release strategies~\citep{glazer2024frontiermath, an2025amo, bytedance_seed_2025_beyondaime, phan2025humanity}. A small number withhold problems or answers behind access controls to reduce leakage~\citep{phan2025humanity, schmitt2025improofbench, ma2026eternalmath, abouzaid2026first}, thereby improving longevity at the cost of transparency and reproducibility.

\section{Data Collection}

This section describes how \shfull{} was assembled, covering the primary contributor pool and submission terms (\S\ref{sec:contributor_details}), the multi-stage collection and filtering pipeline (\S\ref{sec:filtering}), primary-system contributor interviews on creation strategy, the separate ScienceBench bulk-purchase~\citep{sciencebench2025} and the construction of the Refusal split. Further details are deferred to Appendix~\ref{app:datacollection}.

\subsection{Contributor details}
\label{sec:contributor_details}

Across the full collection, 105 contributors provided accepted questions. Of these, 86 came through our primary submission system across 31 organizations, and the remaining 19 came through the ScienceBench~\citep{sciencebench2025} contribution group. Including ScienceBench contributors, the full pool spans 48\% faculty, 23\% graduate students and postdoctoral researchers (3\% master's students and 20\% PhD students or postdocs), 25\% undergraduates, and 5\% with undisclosed affiliation. Among the 86 primary-system contributors, 72 were recruited via direct outreach (emailing mathematics departments and contacting individual PhD students and faculty), while 14 submitted via our website without prior contact. Most accepted questions came from the direct-outreach pool. Primary-system contributors could opt for monetary compensation, authorship on the dataset paper, or both. We allocated a total compensation pool of USD 260{,}000 and paid on a per-accepted-question basis until the quota for each split was filled. Payments were split-dependent (\S\ref{sec:filtering}) and ranged from USD 36 to USD 3{,}623 per question, with a cap of USD 20{,}000 per contributor. Submissions had to be written in English or Korean, typeset in text-only \LaTeX{} (no diagrams or images), and accompanied by a complete solution and an explicit final-answer line. Primary-system contributors were required to sign a submission agreement affirming that each problem had been originally authored without AI assistance. Acceptable subjects span algebra, number theory, combinatorics, analysis, geometry and topology, probability and stochastics, differential equations, and related cross-disciplinary topics. All primary-system contributors signed an NDA and an IP-transfer agreement. The leakage-risk policy and per-gate earnings statistics are recorded in Appendix~\ref{app:contributor_terms}.


\begin{figure}[t] 
\centering 
\includegraphics[width=0.96\textwidth]{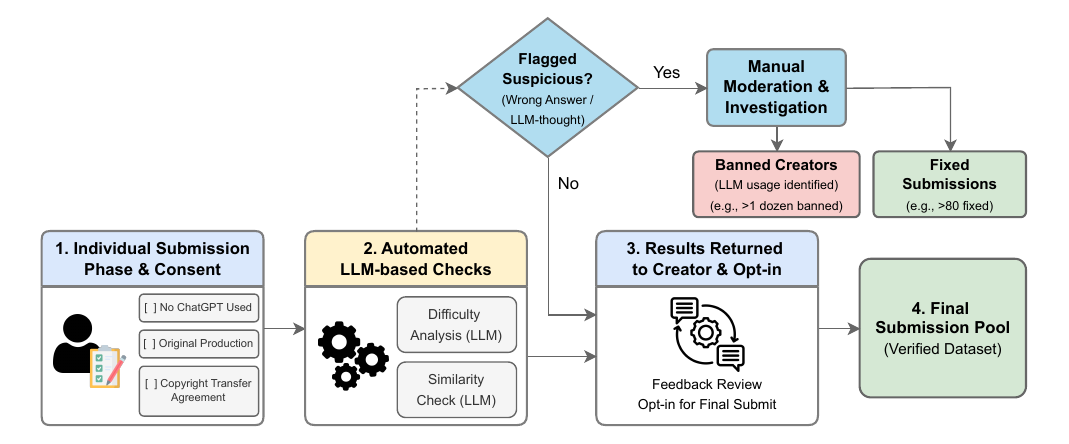} \caption{\footnotesize \textbf{Item-flow through the \gls{soohak} collection pipeline.} Each candidate item passes through submission under an originality and copyright agreement, automated screening with model-gated routing and similarity checks, manual review by two human reviewers, contributor-controlled opt-in, and final inclusion. The figure reports candidate counts at each stage. Banned creators denote contributors found to have submitted AI-generated questions, whose submissions were excluded.}
\label{fig:collection_pipeline} 
\vspace{-3mm}
\end{figure}

\subsection{Collection process} \label{sec:filtering}

\paragraph{Pipeline overview.} Figure~\ref{fig:collection_pipeline} summarizes our five-stage pipeline (submission, automated screening, manual review, contributor-controlled opt-in, and final inclusion). Primary-system contributors upload questions under an agreement affirming original authorship, no AI use, and a copyright grant. LLMs perform automated quality controls; two human reviewers then audit these outputs, and flag suspicious cases, following up with contributors for clarification or revisions when needed. To reduce direct and indirect leakage, only these two reviewers can access pre-opt-in submissions, and a withdrawn or declined submission is immediately deleted so that at most two individuals (often only one) ever viewed it.

\paragraph{Split assignment, manual reviewing, and quality control.}

Each submission through our primary system is first attempted by a panel of baseline LLMs and routed through three model-gated collection gates before final reporting. The first gate requires failure of small open models including \texttt{Qwen3-8B}~\citep{yang2025qwen3} and \texttt{OpenThinker3-7B}~\citep{guha2025openthoughts}. The second gate requires failure of mid-size open models including \texttt{gpt-oss-20B}~\citep{agarwal2025gpt} and \texttt{Qwen3-32B}. The third gate requires failure of all large open models in the panel including \texttt{gpt-oss-120B}, \texttt{Qwen3-235B}, and \texttt{DeepSeek-R1}~\citep{guo2025deepseek}. Questions that pass the first two gates are merged into \shmini{}. The third gate contributes to \shfull{} Challenge, which targets graduate level and research-adjacent mathematics. For \shfull{} Challenge, submission was limited to selected faculty members, postdocs, PhD students, and a small number of IMO medalists in the primary system, and was additionally supplemented with bulk-purchased problems from ScienceBench~\citep{sciencebench2025}. Two human reviewers then audit model-generated solutions against contributor-written references and request clarifications when they disagree. Through this process, we corrected 87 items and banned contributors attempting to submit LLM-generated questions. Additional contribution workflows are described in Appendix~\ref{app:bulk_purchases}. About \(92\%\) of the collected items were originally authored in English, and we translate every item into the other language using a machine-translation-plus-professional-post-editing workflow with LaTeX-preserving placeholders, glossary-normalized mathematical terminology, and an independent QA pass. See Appendix~\ref{app:filtering_full} for full procedural details and Appendix~\ref{app:translation} for the translation workflow.

\subsection{Contributor Interviews}

\begin{tcolorbox}[
  enhanced,
  colback=white,
  colframe=gray!60,
  colbacktitle=gray!25,
  coltitle=black,
  fonttitle=\bfseries,
  title={(\shfull{} Challenge item) Embedding number of a Brieskorn homology 3-sphere},
  boxrule=0.8pt,
  arc=2mm,
  left=3mm,right=3mm,top=2mm,bottom=2mm,
  label={box:brieskorn-embedding}
]
It is known that every closed orientable \(3\)-manifold smoothly embeds in
\(\#_{i=1}^{n}(S^{2}\times S^{2})\) for some \(n\ge 0\).
For a closed orientable \(3\)-manifold \(M\), define \(c(M)\) to be the smallest \(n\ge 0\)
such that \(M\) admits a smooth embedding into \(\#_{i=1}^{n}(S^{2}\times S^{2})\).
Let \(\Sigma(3,5,49)\) be the Brieskorn homology \(3\)-sphere
\[
\Sigma(3,5,49)
=\Bigl\{(z_{0},z_{1},z_{2})\in\mathbb{C}^{3}:\,
|z_{0}|^{2}+|z_{1}|^{2}+|z_{2}|^{2}=1,\ \ z_{0}^{3}+z_{1}^{5}+z_{2}^{49}=0\Bigr\}.
\]
Compute \(c\!\bigl(\Sigma(3,5,49)\bigr)\).

\vspace{1.2mm}
\begin{tcolorbox}[
  enhanced,
  colback=gray!6,
  colframe=gray!35,
  boxrule=0.6pt,
  arc=1.5mm,
  left=2.5mm,right=2.5mm,top=1.5mm,bottom=1.5mm
]
{\small\textbf{Contributor feedback (edited excerpt).}
In my view, the exact value is difficult to infer by merely combining results stated in the existing literature. However, examining the known proof that \(\Sigma(3,5,49)\) bounds an integral homology \(4\)-ball, careful use of Kirby calculus may strengthen the construction, showing that the homology \(4\)-ball is contractible and admits a Mazur-type handle structure. To my knowledge this strengthening is unpublished, and solving the problem likely requires effective Kirby-diagram manipulation.}
\end{tcolorbox}
\end{tcolorbox}

We interviewed data contributors, particularly those who contributed large numbers of items, to understand how the different collection gates were created in practice. A recurring pattern was that \shmini{} items could be written much faster, while a single \shfull{} Challenge problem often required one or more days of work. This reinforces the intended positioning of the benchmark family. \shfull{} Challenge is the part of the dataset that demanded qualitatively more expert effort and produced the strongest headroom against current models.

For \shfull{} Challenge, interviewed primary-system contributors most often described two approaches. First, some submitted research-adjacent questions they had recently been thinking about, where the key step relies on what they termed \emph{folklore-level reasoning}. This means combining standard facts and community heuristics that a professional mathematician could plausibly derive with some work, but that are not packaged as a published theorem. The example in Box~\ref{box:brieskorn-embedding} illustrates this style. Second, contributors often engineered questions from niche research papers. One contributor noted that in an earlier 2025 project separate from ours, a single paper could sometimes be distilled into a hard, self-contained problem. As LLM search and retrieval improved, this became less effective, and creating a \shfull{} Challenge item increasingly required synthesizing ideas across multiple specialized papers. Further interview notes on \shmini{} items, the resulting incentive misalignment, and implied training-data observations are in Appendix~\ref{app:mini_interviews}.

\subsection{Refusal Questions}

\shfull{} Refusal contains items sourced from submissions rejected during quality control because they were ill-posed, including contradictions, missing assumptions, or no unique answer. A model is marked correct on a refusal item only when it diagnoses the flaw instead of confidently producing a numeric answer. Sourcing pool details, prompt criteria, and grading conventions are in Appendix~\ref{app:refusal}.

\subsection{Dataset Details}

\begin{table}[ht]
\centering
\fontsize{8}{10}\selectfont
\caption{Distribution of mathematical subject areas grouped by macro-category.
Individual subject areas are listed with their counts in parentheses. Group
totals are shown in the rightmost column.}
\label{tab:msc_distribution}
\begin{tabular}{p{3cm} p{8.5cm} c}
\toprule
\textbf{Macro-category} & \textbf{Subject areas (count)} & \textbf{Total \#} \\ \midrule

Algebra \& Discrete &
Number theory (269), Combinatorics (131), Algebraic geometry (76), Group theory (67),
Field theory (54), Linear algebra (33), Rings and algebras (24),
Category theory (11), Nonassociative algebra (9), Commutative algebra (6)
& 680 \\ \midrule

Analysis &
Real analysis (115), Series and summability (36), Functional equations (17),
Harmonic analysis (11), Measure theory (9), Complex analysis (9),
Partial differential equations (8), Potential theory (3),
Ordinary differential equations (3), Global analysis (3),
Special functions (2), Functional analysis (1), Operator theory (1),
Calculus of variations (1)
& 233 \\ \midrule

Geometry \& Topology &
Geometry (95), Convex and discrete geometry (30), Algebraic topology (25),
Manifolds (14), Differential geometry (6), General topology (5)
& 175 \\ \midrule

Probability \& Statistics &
Probability theory (24), Statistics (1)
& 25 \\ \midrule

Applied / CS / OR &
Numerical analysis (9), Information and communication theory (7),
Game theory and economics (5), Computer science (4),
Operations research (2)
& 27 \\ \midrule

Logic &
Mathematical logic (1)
& 1 \\ \bottomrule

\end{tabular}
\end{table}

Each question is annotated with two descriptors: (i) \emph{contributor-provided keywords} collected at submission time, and (ii) an \emph{LLM-assigned subject area} used to standardize coverage statistics.

\paragraph{Contributor keywords.}
Problem contributors supplied keyword tags at submission time. Keyword usage mirrors the reporting design, with \shmini{} centered on computational and contest-like pattern finding through tags such as number theory, modular arithmetic, factorization, geometry, and combinatorics. The \challengesplit{} develops a specialized long tail, including tags such as automorphism, abelian variety, Fano variety, Kazhdan--Lusztig polynomials, moduli space, Richardson varieties, Barratt--Eccles operad, and homotopical algebra.

\paragraph{LLM-assigned subjects.}
In addition to keywords, we assign each question to a Mathematics Subject Classification (MSC) subject area using a \texttt{GPT-5-mini} classifier that takes the question plus contributor keywords and maps the item to a fixed taxonomy. The resulting distribution is shown in Table~\ref{tab:msc_distribution}. The dataset is concentrated in \emph{Algebra \& Discrete} (680, driven by number theory (269) and combinatorics (131)), followed by \emph{Analysis} (233), \emph{Geometry \& Topology} (175), with smaller portions in \emph{Applied/CS/OR} (27), \emph{Probability \& Statistics} (25), and \emph{Logic} (1).

\section{Language Model Evaluation Details}

\paragraph{Models.}
We evaluate eleven language models spanning closed and open-weight systems. The closed systems are Gemini-3-Pro~\citep{googledeepmind_gemini_pro}, Gemini-3-Flash, GPT-5 Medium~\citep{singh2025openai}\footnote{blue{We evaluate with GPT-5.1, GPT-5.2, and GPT-5 using identical configurations. GPT-5 yielded the best performance, and thus we report its results in the table.}}, GPT-5-Mini Medium~\citep{singh2025openai}, Claude-Opus-4.5~\citep{anthropic_claude_opus_4_5}, Claude-Sonnet-4.5, and Grok-4.1-Fast. The open-weight systems are Qwen3-235B-A22B-thinking-2507~\citep{yang2025qwen3}, GPT-OSS-120B~\citep{agarwal2025gpt}, Kimi-2.5~\citep{team2026kimi}, and GLM-5~\citep{zai_glm_5_1}. Reasoning was enabled for all models. Reasoning-effort selections, ablation panels, and per-model decoding parameters are available in Appendix~\ref{app:gen_config}.


\paragraph{Sampling and metrics.}
For each model--question pair, we sample three independent responses and report \textbf{avg@3} and \textbf{pass@3}. Let $c_{i,j}\in\{0,1\}$ indicate correctness for question $i$ and sample $j\in\{1,2,3\}$, with $N$ total questions:
\[
\text{avg@3}=\frac{1}{N}\sum_{i=1}^{N}\left(\frac{1}{3}\sum_{j=1}^{3} c_{i,j}\right),\qquad
\text{pass@3}=\frac{1}{N}\sum_{i=1}^{N}\mathbb{I}\Big[\max_{j} c_{i,j}=1\Big].
\]

\paragraph{Answer parsing and judging.}
We parse a final answer from each response. To account for equivalent answer forms, we use GPT-5-Mini as an LLM judge that compares the parsed answer to the gold answer via mathematical equivalence. The judge receives \emph{only} the gold answer and the parsed answer (no question text or solution) and outputs a binary correctness label.

\section{Results}


{\color{blue}
\begin{table}[t]
\fontsize{7.4}{8.4}\selectfont
\centering
\caption{\textbf{Avg@3 and Pass@3 (\%) for SOOHAK-Mini and SOOHAK.} SOOHAK-Mini merges the first two internal model gates. SOOHAK contains Challenge and Refusal. Best per column within each panel is \textbf{bold}. Second best is \underline{underlined}.}
\resizebox{\textwidth}{!}{%
\begin{tabular}{lcccccc}
\toprule
\textbf{Model} &
\multicolumn{2}{c}{\textbf{SOOHAK-Mini} ($n=702$)} &
\multicolumn{4}{c}{\textbf{SOOHAK}} \\
\cmidrule(lr){2-3}\cmidrule(lr){4-7}
& \multicolumn{2}{c}{}
& \multicolumn{2}{c}{\textbf{Challenge} ($n=340$)}
& \multicolumn{2}{c}{\textbf{Refusal} ($n=99$)} \\
\cmidrule(lr){4-5}\cmidrule(lr){6-7}
& \textbf{Avg@3} & \textbf{Pass@3}
& \textbf{Avg@3} & \textbf{Pass@3}
& \textbf{Avg@3} & \textbf{Pass@3} \\
\midrule
\multicolumn{7}{l}{\textit{Closed frontier}} \\
\midrule
\texttt{Gemini-3-Pro} & \underline{71.70} & \underline{80.63} & \textbf{30.39} & \textbf{44.12} & 41.41 & 50.51 \\
\texttt{Gemini-3-Flash} & 61.40 & 68.80 & 15.69 & 25.59 & \textbf{43.10} & 50.51 \\
\texttt{GPT-5} & \textbf{72.22} & \textbf{81.48} & \underline{26.37} & \underline{40.88} & \underline{43.09} & \textbf{61.62} \\
\texttt{GPT-5-Mini} & 67.14 & 78.49 & 18.82 & 28.82 & 41.08 & \underline{55.56} \\
\texttt{Claude-Opus-4.5} & 51.38 & 61.40 & 10.39 & 18.82 & 26.60 & 33.33 \\
\texttt{Claude-Sonnet-4.5} & 40.88 & 51.57 & 5.69 & 10.29 & 27.61 & 35.35 \\
\texttt{Grok-4.1-Fast} & 70.66 & 77.92 & 18.43 & 30.88 & 35.35 & 47.47 \\
\midrule
\multicolumn{7}{l}{\textit{Open-weight (largest in each family)}} \\
\midrule
\texttt{Qwen3-235B-A22B-thinking-2507} & 56.22 & 67.66 & 8.04 & 15.00 & 2.69 & 5.05 \\
\texttt{GPT-OSS-120B} & 61.02 & \textbf{75.21} & \underline{11.27} & \underline{18.53} & \underline{43.77} & \underline{60.61} \\
\texttt{Kimi-2.5} & \textbf{66.07} & \underline{74.93} & \textbf{13.87} & \textbf{20.00} & 29.97 & 42.42 \\
\texttt{GLM-5} & \underline{63.11} & 71.08 & 9.61 & 18.24 & \textbf{49.49} & \textbf{73.74} \\
\bottomrule
\end{tabular}%
}
\label{tab:avg3_pass3_subsets_percent}
\end{table}

}

\paragraph{Overall Performance.} {Table~\ref{tab:avg3_pass3_subsets_percent} reports that scores fall steeply from \shmini{} to \shfull{} Challenge. \texttt{GPT-5} reaches the strongest \shmini{} Avg@3 at 72.22, followed by \texttt{Gemini-3-Pro} at 71.70. \texttt{Gemini-3-Pro} leads Challenge with Avg@3 of 30.39, followed by \texttt{GPT-5} at 26.37. \texttt{GLM-5} leads Refusal with Avg@3 of 49.49. The benchmark family leaves 124 Challenge items unsolved by any evaluated model and 170 items unsolved or missed in total.\footnote{This exceeds smaller benchmarks such as Riemann-Bench ($\geq 23/25$)~\citep{garre2026riemann}. Challenge could have been gated against top closed systems for lower per-item scores, but we chose to preserve scale and the operational collection ladder.}

\paragraph{Open-weight models remain competitive on \shmini{} but trail on \shfull{} Challenge.} {In the lower panel of Table~\ref{tab:avg3_pass3_subsets_percent}, the strongest open-weight systems reach \shmini{} Avg@3 of 66.07 for \texttt{Kimi-2.5} and 63.11 for \texttt{GLM-5}, compared with 72.22 for \texttt{GPT-5} and 71.70 for \texttt{Gemini-3-Pro}. On \shfull{} Challenge, the best open-weight score is 13.87 for \texttt{Kimi-2.5}, compared with 30.39 for \texttt{Gemini-3-Pro} and 26.37 for \texttt{GPT-5}. This gap suggests that open-weight systems transfer less reliably to unpublished and research-adjacent mathematics, which is consistent with recent attempts to apply LLMs to unresolved mathematical problems relying on top-performing closed systems~\citep{alexeev2026short, alexeev2026short2, feng2026semi}. \shfull{} Refusal reverses the ranking. \texttt{GLM-5} reaches the highest Refusal Avg@3 in our evaluation at 49.49, exceeding every closed model, including \texttt{Gemini-3-Flash} at 43.10 and \texttt{GPT-5} at 43.09. The \texttt{Qwen3} family is a clear outlier in the other direction, performing worst on \shfull{} Refusal across the panel.


\paragraph{MSC subfield performance.}
A per-MSC accuracy breakdown across all 18 full-coverage models shows that the per-subfield leader rotates by mathematical flavor. \texttt{Gemini-3-Pro} tops algebra, number theory, and analysis subfields. \texttt{Grok-4.1-Fast} tops the geometric and stochastic subfields. \texttt{GPT-OSS-120B} with hard reasoning and 81920 context tops MSC~15, the only subfield where an open-weight model wins. The full table including uniformly hard or easy subfields and high-disagreement subfields is in Appendix~\ref{app:msc_subfield} and Table~\ref{tab:msc-trends-filled}.

\begin{figure}[t]
\centering
\includegraphics[width=0.98\textwidth]{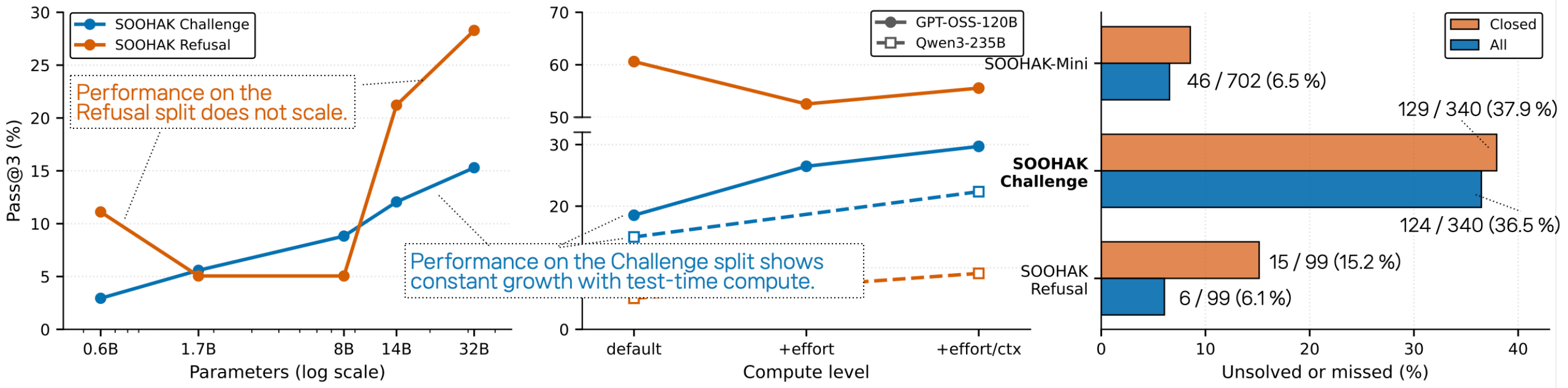}
\caption{\footnotesize \textbf{Compute scaling on Challenge and Refusal and unsolved counts}
\textit{Left:} Pass@3 across the Qwen3 family (0.6B to 32B) on Challenge (blue) and Refusal (orange);
\textit{Middle:} Test-time scaling on the same two splits for GPT-OSS-120B (solid) at three settings (medium-reasoning at 16,384 tokens, hard-reasoning at 16,384 tokens, and hard-reasoning at 81,920 tokens) and for Qwen3-235B-A22B-thinking-2507 (dashed) at two settings (default and 81,920-token context; the model has no exposed reasoning-effort parameter).
\textit{Right:} Fraction of items unsolved by any model in the panel.}
\label{fig:scaling_compute}
\end{figure}

\textbf{Challenge scales roughly linearly with both train- and test-time compute; Refusal does not.}
Within the Qwen3 family (Table~\ref{app:test_time_scaling}, Figure~\ref{fig:scaling_compute}, left), Challenge climbs from 2.94 at 0.6B to 15.29 at 32B, adding roughly 3 points per checkpoint after the first jump. On the test-time axis, raising the per-question token budget produces a comparable trajectory: all models are evaluated at a default 16,384-token budget, and the 81,920-token variant in Table~\ref{app:test_time_scaling} (Figure~\ref{fig:scaling_compute}, right) is a \(5\times\) extension. Under this extension, GPT-OSS-120B (medium \(\rightarrow\) hard \(\rightarrow\) hard with 81,920 tokens) lifts Challenge from 18.53 to 26.47 to 29.71, and Qwen3-235B-A22B-thinking-2507 lifts from 15.00 to 22.35. Since Challenge has not yet saturated on either axis, larger or longer-budget systems (e.g., GPT-5-Pro) would likely raise these numbers further; we omit such runs due to API and compute budget. Notably, Refusal does not show the same scaling patterns; what governs refusal and hallucination behavior we leave to future work.

\section{Human Baselines}

When introducing a challenging benchmark, providing human baselines is important so that future language models have an interpretable reference point for difficulty. 
Note that direct comparison between human and model scores should be interpreted with care, as humans operate under wall-clock time constraints while models are evaluated under token budgets that are not directly comparable.
Because the benchmark family spans olympiad, graduate-course, and research-level problems, a single expert profile is insufficient to characterize what it measures. We therefore assembled five teams with varying mathematical backgrounds. The teams' experience ranged from IMO medalists to published math researchers, so that performance differences across groups can reveal which types of mathematical expertise the benchmark rewards and to what degree.}
We describe each team's profile, and per-team narratives, full credentials, and collaboration strategies in Appendix~\ref{app:human_teams}.



\begin{table}[h]
\centering
\caption{Human participant team profiles.}
\label{tab:human_teams}
\footnotesize
\begin{tabular}{lllp{8cm}}
\toprule
\textbf{Team} & \textbf{Name} & \textbf{Size} & \textbf{Key Credentials} \\
\midrule
A & CS Major (IMO exp.) & 5 & 7 IMO HM, 2 IMO Bronze, 1 EGMO Bronze, 1 APMO Bronze \\
B & Math Major (IMO exp.) & 5 & 1 IMO Silver, 1 APMO Bronze, 1 WMTC Gold, 2 KMS Gold/Silver \\
C & Math Major (IMO Gold) & 5 & 2 IMO Gold, 1 IMO Silver, 2 ICPC (Seoul) Gold, 2 KMS Gold \\
D & Math Major & 5 & 1 KMS Silver, 2 KOI Bronze, 1 ICPC (Asia Pacific) Bronze, SIMC 2.0 Champion \\
E & Math Researchers & 5 & PhD holders in mathematics and computer science \\
\bottomrule
\end{tabular}
\end{table}

\subsection{Evaluation Setup}

We evaluate both human participants and LLMs on the same set of 79 prompts, sampled across benchmark splits. The evaluation set includes 49 Calibration problems and 30 Challenge problems.
We intentionally upsample harder questions. The Challenge items are drawn from narrower subfields and exhibit higher variance even among strong solvers, so a small number of hard items is often insufficient to reliably differentiate PhD-level performance.

Human evaluations were conducted under a nominal 4.5 hour time budget. Participants were permitted to use any non-AI tools, including programming environments, computer algebra systems, and internet search for reference material. We explicitly prohibited the use of LLMs and asked participants to avoid AI-assisted search features to prevent inadvertent model assistance. Session conditions were not fully standardized across sessions. We report this for transparency, as such operational differences can introduce additional variance in measured performance at frontier difficulty. Participants were compensated \$340.

LLMs are evaluated on the same sampled prompts at Pass@1. All reported scores for both humans and LLMs reflect performance on this evaluation set instead of the full dataset. Scoring is purely outcome-based. A prompt is counted as correct only if the final answer is correct. We do not award partial credit for attempted-but-incorrect or partially correct solutions.


\begin{figure}[t]
\centering
\includegraphics[width=0.96\textwidth]{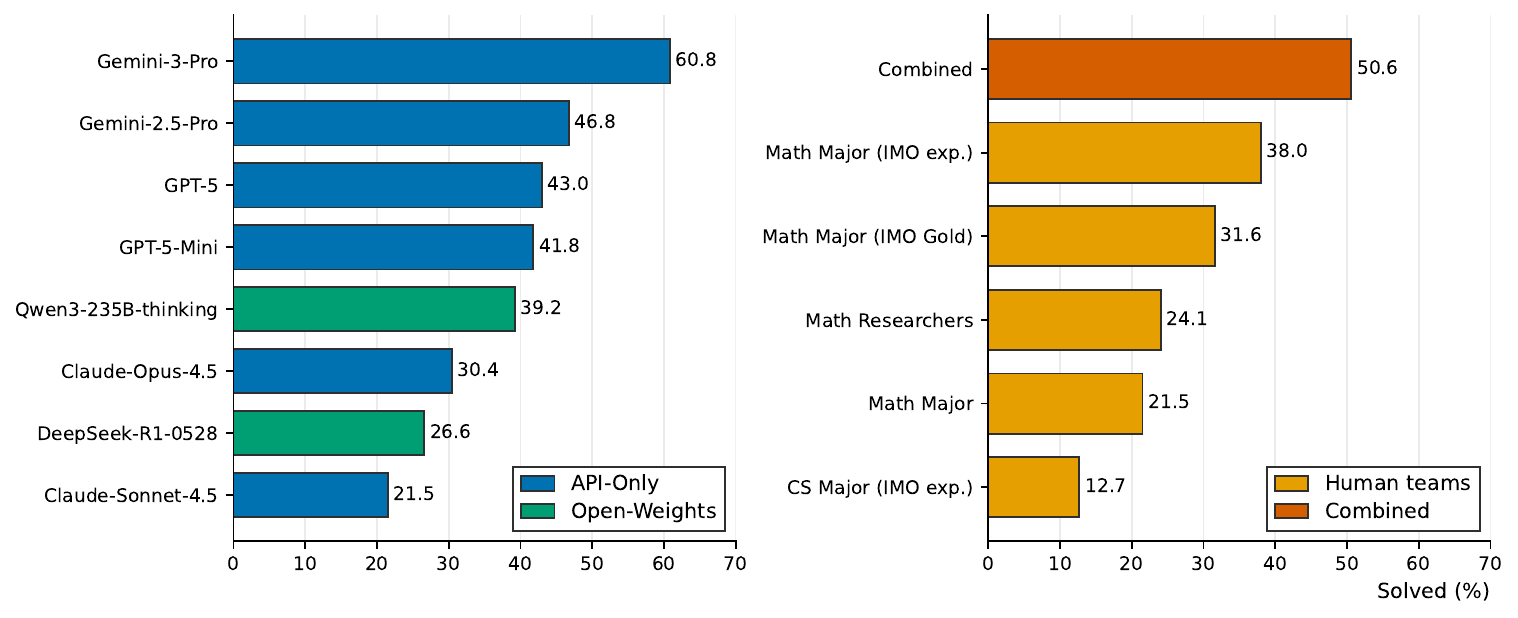} \caption{\footnotesize \textbf{Model and human-team accuracy on the 79-problem human-evaluation set.} The left panel shows closed and open-weight models. The right panel shows individual human teams A through E plus their combined coverage. Only Gemini-3-Pro exceeds combined-human coverage at 50.6\%. The strongest single team is Math Major with IMO experience.}
\label{fig:ai_human_split}
\vspace{-3mm}
\end{figure}

\subsection{Evaluation Results}

\paragraph{Human coverage via aggregation.}
\emph{Combined} denotes the \emph{union} of all questions solved by any human team. A question is counted as solved if at least one team solved it. \texttt{Gemini-3-Pro} at 60.8\% is the only model that exceeds combined human coverage at 50.6\%. No single human profile fully covers the benchmark's breadth, but collectively the five groups provide good coverage across benchmark splits. At the single-team level, \textbf{Math Major with IMO experience} is the strongest human team, yet multiple models surpass it.

\paragraph{Contest training, not research experience, drives performance.}
The performance ordering across groups reveals that the benchmark primarily rewards contest-style mathematical reasoning under time pressure, not research-level expertise. Teams with sustained olympiad experience combined with undergraduate mathematics training, Math Major with IMO experience and Math Major with IMO Gold, achieve the highest scores among human participants. Math Researchers, despite having the deepest mathematical expertise, score lower than the top undergraduate teams. We view this as a \emph{task-format mismatch}, not an ability gap. Two factors likely drive this. The 4.5-hour budget incentivizes short-path solutions more natural to contest-trained mathematicians, and the benchmark's breadth limits the advantage of narrow research specialization. CS Major with IMO experience underperforms the pure-math teams despite comparable olympiad credentials, further suggesting that sustained undergraduate-level pure-math training provides an additional advantage on this benchmark.

\paragraph{Tools and process matter.}
Within the undergraduate teams, more active use of computational tools for calculation and verification qualitatively correlates with better outcomes, consistent with the importance of speed and error control under time pressure. Beyond skill, we observe substantial behavioral and organizational effects. Engagement appears higher in shared-room settings than in isolated settings, and collaboration strategy appears to shift the throughput-accuracy trade-off in meaningful ways. Math Major with IMO Gold emphasized parallelism, while Math Major with IMO experience emphasized division of labor with cross-checking. Finally, humans tend to avoid long, notation-heavy items even when they are not intrinsically difficult, whereas LLMs apply more uniform effort across question lengths, yielding an additional coverage advantage beyond raw competence.

\section{Conclusion}
\label{sec:discussion}

We introduce \shfull{}, a benchmark for evaluating graduate level and research-adjacent mathematical reasoning. \shfull{} consists of a 340-item Challenge subset and a 99-item Refusal subset. The Challenge subset is newly authored by expert contributors and remains difficult for frontier systems, with the best model reaching only 30.39\% Avg@3. Open-weight systems show a sharper drop, with the strongest open-weight model reaching 13.87\% Avg@3 on Challenge. The Refusal subset exposes a complementary failure mode, since even strong closed models often attempt to answer prompts that are invalid as written. To support broader tracking, we also release \shmini{}, a 702-item companion subset spanning high-school olympiad through early graduate material. Together, \shfull{} and \shmini{} form a contamination-resistant full collection created by 105 contributors across two collection channels. The full collection will be open-sourced in late 2026.


\newpage



\section*{Acknowledgments}


We thank the following dataset contributors: Christos Athanasiadis (National and Kapodistrian University of Athens), François Glineur (UCLouvain), In-Jee Jeong (KAIST), Jehan Oh (Kyungpook National University), Sam Hopkins (Howard University), Seunghyun Yoon (Seoul National University), and Youngmin Lee (Kyonggi University).

We would also like to thank EleutherAI and CoreWeave for providing compute for some of the evaluations.

\bibliographystyle{apalike}
\bibliography{references}


\newpage
\appendix

\newcommand{\apptocentry}[2]{%
  \noindent\textbf{\ref{#1}\quad\nameref{#1}}%
  \leaders\hbox to 0.6em{\hss.\hss}\hfill%
  \textbf{\pageref{#1}}\par
}

\newcommand{\apptocsubentry}[2]{%
  \noindent\hspace{1.8em}%
  \ref{#1}\quad\nameref{#1}%
  \leaders\hbox to 0.6em{\hss.\hss}\hfill%
  \pageref{#1}\par
}

\newcommand{\apptocsubentryclaude}[2]{%
  \noindent\hspace{1.8em}%
  \claude{\ref{#1}\quad\nameref{#1}}%
  \leaders\hbox to 0.6em{\hss.\hss}\hfill%
  \pageref{#1}\par
}

\begingroup
\small
\setlength{\parindent}{0pt}
\setlength{\parskip}{4pt}

\apptocentry{app:affiliations}{}
\apptocentry{app:datacollection}{}

\apptocsubentry{app:funding}{}
\apptocsubentry{app:contributor_terms}{}
\apptocsubentry{app:filtering_full}{}
\apptocsubentry{app:bulk_purchases}{}
\apptocsubentry{app:mini_interviews}{}
\apptocsubentry{app:problem_type_composition}{}
\apptocsubentry{app:translation}{}
\apptocsubentry{app:refusal}{}

\apptocentry{app:examples}{}
\apptocentry{app:eval_extensions}{}

\apptocsubentry{app:gen_config}{}
\apptocsubentry{app:qwen_scaling}{}
\apptocsubentry{app:test_time_scaling}{}
\apptocsubentry{app:carefulness}{}
\apptocsubentry{app:msc_subfield}{}

\apptocentry{app:human_teams}{}

\endgroup


\section{Author affiliations}
\label{app:affiliations}

The numbered superscripts in the Organizing Team and Dataset Contributors blocks on the front page correspond to the institutions below.

\begin{flushleft}
{$^{1}$OneLineAI,
$^{2}$EleutherAI,
$^{3}$Carnegie Mellon University,
$^{4}$Dnotitia Inc.,
$^{5}$Saltlux Innovation,
$^{6}$SYSTRAN Korea,
$^{7}$Rajah Serfoji Government College,
$^{8}$Ruhr-University Bochum,
$^{9}$DeltaX,
$^{10}$Sungkyunkwan University,
$^{11}$Far Eastern Federal University, School of Economics and Management,
$^{12}$Korea Advanced Institute of Science and Technology (KAIST),
$^{13}$Chonnam National University,
$^{14}$Chungnam National University,
$^{15}$Institute of Informatics and Telematics (IIT), CNR,
$^{16}$USC,
$^{17}$Hotam and PV Lyceum,
$^{18}$Sejong University,
$^{19}$UNC Chapel Hill,
$^{20}$Ewha Womans University,
$^{21}$Seoul National University,
$^{22}$Sogang University,
$^{23}$Dongguk University - Seoul Campus,
$^{24}$TU Dresden,
$^{25}$Arizona State University,
$^{26}$Chung-Ang University,
$^{27}$Universiti Malaya,
$^{28}$Warsaw School of Economics,
$^{29}$Inria,
$^{30}$Hankuk University of Foreign Studies,
$^{31}$Independent,
$^{32}$Private University College of Education of the Diocese of Linz,
$^{33}$DMTLABS,
$^{34}$National Information Society Agency (NIA),
$^{35}$The University of Seoul,
$^{36}$University of Chile.
}
\end{flushleft}


\section{Data collection details}
\label{app:datacollection}

This appendix collects operational details for the data-collection pipeline summarized in body~\S3. It covers the funding context, primary-system contributor terms, routing and review, the ScienceBench bulk-purchase protocol, contributor reports on \shmini{} and Challenge creation strategies, the bilingual translation pipeline, and the construction and grading of the Refusal split. Throughout this appendix, \shfull{} refers to Challenge plus Refusal. The full collection refers to \shfull{} together with the companion \shmini{} subset.

\subsection{Funding context and Sovereign-AI background}
\label{app:funding}

The full collection was developed for the South Korean Ministry of Science and ICT MSIT ``Sovereign AI Foundation Model'' project, which launched an elite-team competition in August 2025~\citep{msit2025_fullscale_launch}. The benchmark was designed as part of an elimination stage of that competition, which required an evaluation that is fair across model families, resistant to contamination, and stable enough to distinguish systems across multiple rounds. To meet these requirements on a compressed schedule, the end-to-end effort runs from August to December 2025, spanning contributor recruitment, dataset collection, manual review, and the initial evaluation. The project is supported by approximately \$550{,}000 USD or KRW 800 million in government funding, with a contractual mandate for strict confidentiality and rigorous methodology. The temporary embargo and late-2026 release date reflect this funding context. The embargo is bounded by the conclusion of the funded evaluation initiative. In the interim, the project organizers welcome requests for model evaluation, and we will provide periodic updates to this paper to reflect results on newly released global models.

\subsection{Contributor terms, compensation eligibility, and submission statistics}
\label{app:contributor_terms}

This subsection records the operational terms governing primary-system contributor participation, deferred from body~\S3.1, together with submission and earnings statistics.

\paragraph{Compensation eligibility, confidentiality, and IP transfer.}
Compensation was issued only after a primary-system submission passed our vetting pipeline and its final split assignment was confirmed. Prior to payment eligibility, primary-system contributors were required to execute confidentiality and IP-transfer documents (NDA and work-made-for-hire or copyright assignment). Upon acceptance, all rights to the problem statement, solution, metadata, and source files were transferred to the project, and the authors were not allowed to reuse, republish, distribute, or create derivative versions of accepted material. To reduce leakage risk, primary-system contributors were instructed not to paste candidate questions into public chat interfaces (e.g., ChatGPT) and to use the designated submission workflow.

\paragraph{Guideline materials and participant instructions.}
The public project page and full guideline document used for participant-facing instructions are available at \href{https://novamath.github.io/}{\texttt{novamath.github.io}} and \href{https://drive.google.com/file/d/1Zl0S-DNZLZQvBb7j7Oe5-VfBmsAJfM_T/view}{the guideline document}, respectively. These materials supplement the operational terms above and document the problem-format requirements, originality and no-AI-use constraints, confidentiality expectations, submission workflow, and compensation structure used during collection.

\paragraph{Submission and earnings statistics.}
Within the primary submission system, a total of 101 mathematicians and students submitted, and 86 had at least one question accepted into the full collection. Among the invited contributors, the top five contributors, all students, submitted 434 questions in total. These questions were mostly routed into the first collection gate and are reported within \shmini{}. The students were invited offline to our office and validated that their questions were not LLM-generated. In contrast, the top five earners received USD 68k in total, with no overlap with the top submission group. These top earners were predominantly faculty members whose contributions were mostly routed into Challenge. Public submissions were smaller in scale. The top three contributors submitted 20 questions in total, and the top three earners received a total of USD 1{,}703. The full collection also includes 19 ScienceBench contributors, described in Appendix~\ref{app:bulk_purchases}, yielding 105 contributors in total.

\subsection{Item filtering: split assignment, manual reviewing, and quality control}
\label{app:filtering_full}

\paragraph{Split assignment.}
We assign each primary-system candidate question using three model-gated collection gates. Each primary-system submission is attempted by a panel of baseline LLMs. Gate 1 requires failure of small open models including \texttt{Qwen3-7B}~\citep{yang2025qwen3} and \texttt{OpenThinker3-7B}~\citep{guha2025openthoughts}. Gate 2 requires failure of mid-size open models including \texttt{gpt-oss-20B}~\citep{agarwal2025gpt} and \texttt{Qwen3-32B}. The Challenge gate requires failure of all large open models in the panel including \texttt{gpt-oss-120B}, \texttt{Qwen3-235B}, and \texttt{DeepSeek-R1}~\citep{guo2025deepseek}. These model-based criteria are enforced as a hard policy. We report Gates 1 and 2 jointly as the companion \shmini{} subset. The main contribution \shfull{} consists of the \challengesplit{} and the Refusal split. As non-binding guidance, \shmini{} often corresponds to high-school through lower-undergraduate material, including olympiad and textbook-like problems. Challenge targets graduate level and research-adjacent material.

\paragraph{Manual reviewing.}~Two members of our team manually audit the automated screening outputs and communicate the results to the original authors by email. Reviewers read the model-generated solutions produced during the difficulty-rating stage and compare them against the provided reference answer/solution. When a model produces a coherent and seemingly correct solution that conflicts with (or casts doubt on) the submitted reference answer, we flag the item and request clarification. Based on the ensuing correspondence, authors may revise and resubmit or opt in to inclusion of the clarified item in the full collection. If no response is received, we exclude the submission. Using this primary-system process, we corrected 87 submissions and banned multiple contributors for attempting to submit LLM-generated questions.

\paragraph{Quality control.}
For each primary-system submission, we perform an automated consistency check by comparing the contributor's proposed answer against answers produced by multiple LLMs generated as part of our split-assignment pipeline. Exact agreement provides supportive evidence of correctness. When the answers disagree, we return the item and the generated response to the author for a delayed re-solve. We do not perform full cross-validation across the full collection, since obtaining independent third-party solutions is often prohibitively difficult for many \shmini{} items and most Challenge items. Instead, an external evaluation organization examined a randomly sampled portion of the collection after the main collection phase and flagged approximately 5\% of items for potential issues. Flagged does not imply incorrect. Flags included genuine errors as well as ambiguity and evaluator misunderstanding. We addressed all flagged cases via correction or clarification and treat the flag rate as a conservative upper bound. Overall, we estimate that the fraction of items with substantive errors is \emph{at most} 5\%. As benchmark difficulty increases, exhaustive answer validation becomes increasingly costly and sometimes infeasible in practice, a challenge encountered in other frontier-level evaluations such as Humanity's Last Exam~\citep{zhai2026hle, skarlinski2025hle_biochem_wrong} and FrontierMath~\citep{burnham2025frontiermath_within_reach}. We report these procedures and the resulting upper bound so that users can interpret results with appropriate caution.

\subsection{Bulk purchases and ScienceBench contribution protocol}
\label{app:bulk_purchases}

Besides the primary collection described above, we acquired an additional set of Challenge problems through a bulk purchase from ScienceBench~\citep{sciencebench2025}. Nineteen additional contributors participated including 10 professors, 7 postdoctoral researchers, and 2 PhD students. This yielded 112 problems that we integrated into Challenge and brought the full collection to 105 contributors in total. Contributors were instructed to write a single prompt framed as a problem or exercise with a unique, well-defined answer, to supply a precise solution with multiple nontrivial reasoning steps, and to ensure that computation was not the primary source of difficulty. Contributors could view other submitted prompts without solutions and were encouraged to attempt solving them and leave comments. This lightweight peer-review process led to revisions and improvements for several items. Because every submission was solved by at least one model, verification focused on confirming that the model solutions judged correct and the contributor-provided solution relied on the same core reasoning ideas. To the best of our knowledge, we did not observe cases where a model produced a correct solution using reasoning that was previously unknown to the author at the time of submission.

\subsection{SOOHAK-Mini and Challenge creation strategies}
\label{app:mini_interviews}

For \shmini{}, a common strategy was to remix existing contest problems, often inspired by sources such as the IMO Shortlist. Primary-system contributors preserved the core invariant or trick while changing parameters, recombining constraints, or reframing the setting. They also reported that, even when such remixes were genuinely new and not direct copies of standard templates, models with stronger upper-math capabilities could often solve them quickly by leveraging broader mathematical knowledge than the intended high-school contest audience. In retrospect, this created an incentive misalignment. Original contest-style problems were time-consuming to craft yet less likely to be rewarded as Challenge items, while paper-based or research-adjacent constructions were viewed as a more reliable path to higher compensation. More broadly, these creation strategies suggest that current LLM weaknesses are strongly shaped by access to training data. When relevant mathematics is absent from accessible papers, scattered across niche sources, or hidden behind paywalls, contributors found it substantially easier to write questions that remain challenging for frontier systems.

\subsection{Problem-type composition}
\label{app:problem_type_composition}

In addition to split assignment, we assign each question a coarse \emph{problem type} label using a \texttt{GPT-5} classifier. The labels are \emph{olympiad} for contest-style problems with short trick-based solutions, \emph{undergrad} for standard university topics such as calculus, linear algebra, and introductory algebra and analysis, \emph{graduate} for graduate core and research-level techniques, and \emph{beyond} for specialized research problems requiring background beyond graduate core. The reporting split sizes are 702 \shmini{} items, 340 Challenge items, and 99 Refusal items. \shmini{} contains 401 olympiad, 218 undergrad, 63 graduate, and 18 beyond items. Challenge contains 77 olympiad and 69 undergrad items because the classifier tracks style and background, not the collection gate. The remaining Challenge items are graduate or beyond items, which explains the stronger shift toward advanced mathematics in \shfull{} Challenge.

\subsection{Translation pipeline}
\label{app:translation}

Approximately \(92\%\) of the items were originally authored in English. To construct a parallel bilingual benchmark, we translated every item into the other language (Korean\(\rightarrow\)English and English\(\rightarrow\)Korean) using a machine-translation-plus-post-editing workflow. Concretely, we first replace all LaTeX spans (expressions, symbols, and commands) with protected placeholders, generate a draft translation with a domain-adapted machine translation system, and then restore the placeholders so that all mathematical content is preserved verbatim.

Draft translations are subsequently post-edited by professional translators under a strict guideline emphasizing semantic faithfulness, terminology consistency, standardized mathematical notation, and minimal paraphrasing. Changes to mathematical symbols or LaTeX are disallowed except when required by target-language grammar. An independent reviewer then performs quality assurance to identify mistranslations, omissions, terminology inconsistencies, punctuation issues, and text--formula mismatches. We also performed automated checks for LaTeX renderability and formula equivalence, and normalized mathematical terminology using a curated glossary based on the Korean Mathematical Society dictionary\footnote{\url{https://www.kms.or.kr/mathdict/list.html}}. Finally, automated quality assurance scripts flag untranslated segments and cross-item term inconsistencies. Final bilingual items are serialized in a structured JSON format. All translation and QA steps were executed under strict security constraints, including prohibitions on uploading full problems to external LLM interfaces or sharing files outside the secured workflow.

\subsection{Refusal question sourcing and grading}
\label{app:refusal}

We include \emph{refusal questions} to measure whether systems can recognize and appropriately respond to ill-posed or unanswerable prompts. At frontier difficulty, a common failure mode is producing a persuasive but incorrect solution when the problem statement is internally inconsistent, underspecified, or otherwise lacks a well-defined answer. Refusal questions are designed to probe this form of benchmark hallucination and overconfidence. Refusal questions are drawn from submissions that we \emph{rejected} during quality control because they exhibited logical flaws, missing assumptions, or other issues that render the question unsolvable or non-unique as stated. Concretely, we maintain a pool of such items and select from this pool to create refusal-control prompts. These items are not included as Challenge questions. For refusal questions, the desired response is a \emph{diagnostic refusal}. The solver should explicitly indicate that the question has no well-defined answer \emph{as written} or that it is underspecified, ideally with a brief explanation of the issue. We mark responses as incorrect if they present a specific mathematical answer as though the item were well-posed, or if they refuse without engaging the mathematical validity of the prompt.


\section{Example problems}
\label{app:examples}

This subsection collects standalone example boxes referenced from the body. The Brieskorn 3-sphere example illustrating folklore-level reasoning on the \challengesplit{} appears in Box~\ref{box:brieskorn-embedding}. The box below shows a \shmini{} item used to illustrate the SOOHAK-Mini creation style discussed in Appendix~\ref{app:mini_interviews}.

\begin{tcolorbox}[
  enhanced,
  colback=white,
  colframe=gray!60,
  colbacktitle=gray!25,
  coltitle=black,
  fonttitle=\bfseries,
  title={(\shmini{} item) Polynomial Functional Equation / Degree Forcing},
  boxrule=0.8pt,
  arc=2mm,
  left=3mm,right=3mm,top=2mm,bottom=2mm,
  label={box:poly-degree-forcing}
]
Let \(f:\mathbb{R}\to\mathbb{R}\) be a polynomial such that for all real \(x\),
\[
f(x+1)=2f(x)-5x
\qquad\text{and}\qquad
f(f(x))=25x+30.
\]
Compute \(f(1)\).

\vspace{1.2mm}
\begin{tcolorbox}[
  enhanced,
  colback=gray!6,
  colframe=gray!35,
  boxrule=0.6pt,
  arc=1.5mm,
  left=2.5mm,right=2.5mm,top=1.5mm,bottom=1.5mm
]
{\small\textbf{Construction notes (edited excerpt).}
This is a degree-forcing functional equation. The shift relation \(f(x+1)=2f(x)-5x\) rules out \(\deg f\ge 2\) by leading-term comparison, so \(f\) must be linear. Then \(f(x)=ax+b\) is pinned by coefficient matching and checked by the composition constraint \(f(f(x))=25x+30\). This question is referenced from the IMO theme ``polynomial functional equations where one relation forces low degree and another determines parameters.''}
\end{tcolorbox}

\end{tcolorbox}


\section{Evaluation extensions}
\label{app:eval_extensions}

This appendix collects evaluation-side extensions referenced from body~\S5 and \S6. It covers per-model decoding settings, a per-parameter Pass@3 view of the \texttt{Qwen3} family, test-time-scaling ablations on extended reasoning effort and context budget, a carefulness-adjusted ranking that penalizes confidently wrong answers on the Refusal split, and the full MSC-subfield breakdown summarized in body~\S6.

\subsection{Generation configuration}
\label{app:gen_config}

For every model we follow the decoding configuration recommended by the provider. Table~\ref{tab:gen_config} lists the temperature, reasoning configuration, and context budget used for each system. Two conventions hold across the panel. \texttt{Gemini}, \texttt{GPT}, and \texttt{GPT-OSS} families are sampled at temperature 1.0, in line with vendor guidance for their reasoning modes. The remaining systems use temperature 0.6, the recommended setting for their thinking variants. Reasoning is enabled wherever it is exposed, including the \texttt{thinking} variant for the \texttt{Qwen3} family, \emph{medium} reasoning effort for \texttt{GPT-5}, \texttt{GPT-5-Mini}, and \texttt{GPT-OSS-120B} by default, and provider-default extended thinking for \texttt{Claude}. Top-$p$ and top-$k$ are left at provider defaults. The samples reported in the main results in Table~\ref{tab:avg3_pass3_subsets_percent} use these settings. The test-time-scaling ablations in Appendix~\ref{app:test_time_scaling} vary the reasoning-effort level and context budget while holding temperature fixed.

\begin{table}[h]
\centering
\small
\caption{\textbf{Per-model decoding configuration used in the main results.} Temperature follows provider-recommended values for the reasoning mode. Top-$p$ and top-$k$ are provider defaults.}
\label{tab:gen_config}
\begin{tabular}{llll}
\toprule
\textbf{Model} & \textbf{Temp.} & \textbf{Reasoning} & \textbf{Context} \\
\midrule
\multicolumn{4}{l}{\textit{Closed}} \\
\texttt{Gemini-3-Pro}                   & 1.0 & enabled (default)        & default \\
\texttt{Gemini-3-Flash}                 & 1.0 & enabled (default)        & default \\
\texttt{GPT-5}                          & 1.0 & medium reasoning         & default \\
\texttt{GPT-5-Mini}                     & 1.0 & medium reasoning         & default \\
\texttt{Claude-Opus-4.5}                & 0.6 & extended thinking        & default \\
\texttt{Claude-Sonnet-4.5}              & 0.6 & extended thinking        & default \\
\texttt{Grok-4.1-Fast}                  & 0.6 & enabled (default)        & default \\
\midrule
\multicolumn{4}{l}{\textit{Open-weight}} \\
\texttt{Qwen3-235B-A22B-thinking-2507}  & 0.6 & thinking variant         & default \\
\texttt{GPT-OSS-120B}                   & 1.0 & medium reasoning         & default \\
\texttt{Kimi-2.5}                       & 0.6 & thinking variant         & default \\
\texttt{GLM-5}                          & 0.6 & thinking variant         & default \\
\bottomrule
\end{tabular}
\end{table}

\subsection[Qwen3 size scaling]{\texttt{Qwen3} size scaling}
\label{app:qwen_scaling}

Table~\ref{tab:qwen3_scaling} reports Pass@3 across the \texttt{Qwen3} family from 0.6B to 235B parameters, separated into the standard release and the \texttt{-2507} thinking-tuned checkpoints. Within the standard panel, \shmini{} Pass@3 scales monotonically from 35.75 at 0.6B to 70.80 at 32B. Challenge Pass@3 rises from 2.94 to 15.29 across the same sequence. Within the \texttt{-2507} panel, the \texttt{235B-A22B-thinking-2507} model regresses on \shmini{} at 67.66 relative to the \texttt{30B-A3B-thinking-2507} model at 71.23. The 81920-token-context variant in Table~\ref{tab:test_time_scaling} partially recovers to 70.23, suggesting that the largest \texttt{Qwen3} model is bottlenecked on output length for this benchmark.

\begin{table}[h]
\centering
\small
\caption{\textbf{\texttt{Qwen3} family Pass@3 by parameter count.} Top panel shows the standard \texttt{Qwen3} release. Bottom panel shows \texttt{-2507} thinking-tuned checkpoints.}
\label{tab:qwen3_scaling}
\begin{tabular}{lccc}
\toprule
\textbf{Model} &
\multicolumn{1}{c}{\textbf{\shmini{}}} &
\multicolumn{2}{c}{\textbf{\shfull{}}} \\
\cmidrule(lr){2-2}\cmidrule(lr){3-4}
& \textbf{Pass@3} & \textbf{Challenge} & \textbf{Refusal} \\
\midrule
\multicolumn{4}{l}{\textit{\texttt{Qwen3} (standard)}} \\
\midrule
\texttt{Qwen3-0.6B}                            & 35.75 &  2.94 & 11.11 \\
\texttt{Qwen3-1.7B}                            & 46.30 &  5.59 &  5.05 \\
\texttt{Qwen3-8B}                              & 60.68 &  8.82 &  5.05 \\
\texttt{Qwen3-14B}                             & 66.38 & 12.06 & 21.21 \\
\texttt{Qwen3-32B}                             & 70.80 & 15.29 & 28.28 \\
\midrule
\multicolumn{4}{l}{\textit{\texttt{Qwen3} \texttt{-2507} thinking}} \\
\midrule
\texttt{Qwen3-4B-thinking-2507}                & 63.25 &  8.24 & 24.24 \\
\texttt{Qwen3-30B-A3B-thinking-2507}           & 71.23 & 15.59 & 21.21 \\
\texttt{Qwen3-235B-A22B-thinking-2507}         & 67.66 & 15.00 &  5.05 \\
\bottomrule
\end{tabular}
\end{table}

\subsection{Test-time scaling with extended context and reasoning effort}
\label{app:test_time_scaling}

Table~\ref{tab:test_time_scaling} reports the effect of increased reasoning effort and extended output budgets on the two open-weight model families with such variants. Extending \texttt{GPT-OSS-120B} from \emph{medium} to \emph{hard} reasoning gains 1.57 points on \shmini{} and 7.94 points on Challenge. Further extending the context budget to 81920 tokens adds another 4.13 points on \shmini{} and 3.24 points on Challenge. \texttt{Qwen3-235B-A22B-thinking-2507} shows the largest Challenge swing from extended context. Pass@3 rises from 15.00 to 22.35.

\begin{table}[h]
\centering
\small
\caption{\textbf{Test-time scaling Pass@3 for SOOHAK-Mini and SOOHAK.} Extended context and reasoning effort improve Challenge more strongly than \shmini{}.}
\label{tab:test_time_scaling}
\begin{tabular}{lccc}
\toprule
\textbf{Variant} &
\multicolumn{1}{c}{\textbf{\shmini{}}} &
\multicolumn{2}{c}{\textbf{\shfull{}}} \\
\cmidrule(lr){2-2}\cmidrule(lr){3-4}
& \textbf{Pass@3} & \textbf{Challenge} & \textbf{Refusal} \\
\midrule
\texttt{Qwen3-235B-A22B-thinking-2507} (default ctx)   & 67.66 & 15.00 &  5.05 \\
\texttt{Qwen3-235B-A22B-thinking-2507} (81920 ctx)     & 70.23 & 22.35 &  9.09 \\
\midrule
\texttt{GPT-OSS-120B} (\emph{medium} reasoning)        & 75.21 & 18.53 & 60.61 \\
\texttt{GPT-OSS-120B} (\emph{hard} reasoning)          & 76.78 & 26.47 & 52.53 \\
\texttt{GPT-OSS-120B} (\emph{hard}, 81920 ctx)         & 80.91 & 29.71 & 55.56 \\
\bottomrule
\end{tabular}
\end{table}

\subsection{Carefulness-adjusted ranking}
\label{app:carefulness}

The Refusal split is informative on its own, but it also interacts with the reasoning splits in a way that the per-split Pass@3 numbers in Table~\ref{tab:avg3_pass3_subsets_percent} make hard to read at a glance. A model that scores well on \shmini{} and Challenge but poorly on Refusal is, in effect, confidently wrong on ill-posed prompts. To surface this, we define three composite scores from the per-split Pass@3.
\begin{align*}
\text{Capability}    &= \tfrac{1}{2}\bigl(\text{SOOHAK-Mini} + \text{Challenge}\bigr), \\
\text{Avg-R}         &= \tfrac{1}{3}\bigl(\text{SOOHAK-Mini} + \text{Challenge} + \text{Refusal}\bigr), \\
\text{SOOHAK-R}      &= \tfrac{1}{2}\bigl(\text{Challenge} + \text{Refusal}\bigr).
\end{align*}
\emph{Capability} is the unweighted mean of the two reasoning splits and contains no carefulness signal. \emph{Avg-R} folds Refusal in as an equally weighted third dimension, so a low Refusal Pass@3 pulls the score down even when reasoning is strong. \emph{SOOHAK-R} is a frontier-focused composite that pairs the hardest reasoning split with Refusal. The metric is purely descriptive, but it has an interpretation as a penalty. Writing $\text{Avg-R} = \text{Capability} - \tfrac{1}{3}(\text{Capability} - \text{Refusal})$ shows that Avg-R subtracts a third of the gap between reasoning capability and refusal performance from raw Capability.

\begin{figure}[h]
\centering
\includegraphics[width=0.96\textwidth]{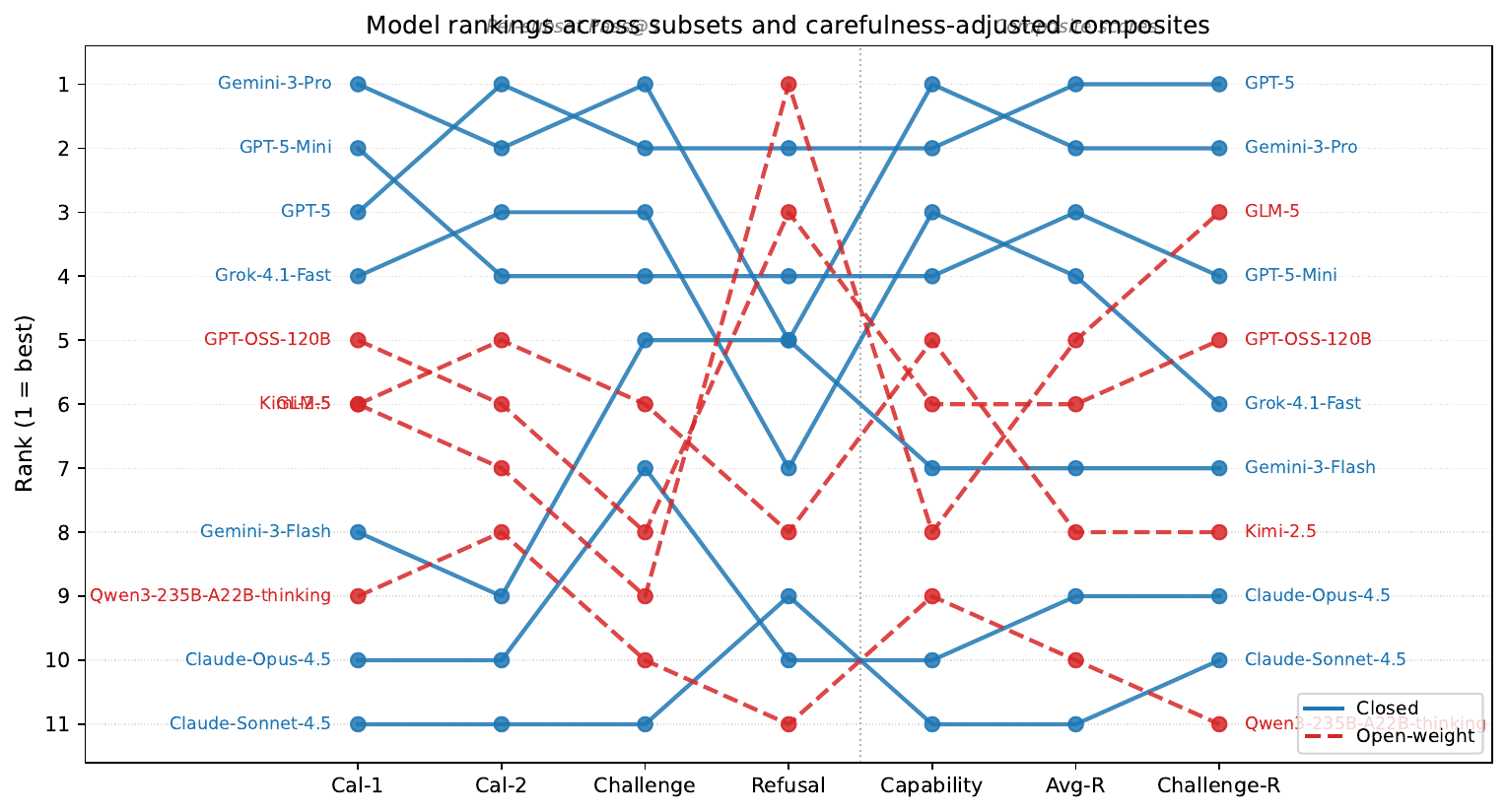}
\caption{\textbf{Model rankings across per-subset Pass@3 and the three composite scores.} Lower is better, with rank 1 at top. To the right of the dotted separator, models that are good at reasoning but careless on Refusal drop in rank. Models that are careful but mid-capability rise. \texttt{GLM-5} rises 3 ranks from Capability to Avg-R. \texttt{Kimi-2.5} drops 3 ranks. \texttt{GPT-5} takes the top Avg-R rank from \texttt{Gemini-3-Pro} despite \texttt{Gemini}'s higher Capability.}
\label{fig:carefulness_rankings}
\end{figure}

\subsection{MSC subfield breakdown}
\label{app:msc_subfield}

To localize model weaknesses, we evaluate accuracy by MSC subject area across all 18 closed and open-weight full-coverage models, restricting to MSCs with at least $n \ge 20$ problems. Models are strongest in MSC~40 (mean $69.8\%$, best $80.6\%$), MSC~26 (mean $59.4\%$, best $75.1\%$), and MSC~11 (mean $57.8\%$, best $71.9\%$), and weakest in MSC~16 (mean $14.6\%$, best $48.6\%$) and MSC~52 (mean $24.7\%$, best $57.8\%$). These uniformly high and low patterns likely reflect subfield difficulty. In contrast, the most diagnostic subfields are those with large across-model gaps. MSC~52 has a range of $57.8$ pp, best score of $57.8\%$, and mean of $24.7\%$. MSC~60 has a range of $55.6$ pp, best score of $59.7\%$, and mean of $45.7\%$. MSC~40 has a range of $52.8$ pp, best score of $80.6\%$, and mean of $69.8\%$. These gaps indicate meaningful differences in subfield-specific competence beyond overall dataset difficulty. The leading model varies by subfield. \texttt{Gemini-3-Pro} leads number-theoretic MSC~11, analysis-oriented MSC~26, combinatorial MSC~5, and algebraic MSC~16. \texttt{Grok-4.1-Fast} leads MSC~40 (series/summability), MSC~51 (geometry), and MSC~60 (probability). \texttt{GPT-OSS-120B} (hard, 81920 ctx) leads MSC~15 (linear algebra), the first MSC in our evaluation where an open-weight system tops the per-subfield ranking.

\begin{table}[h]
  \centering
\fontsize{7.4}{8.4}\selectfont
    \caption{\textbf{MSC trend highlights across 18 closed and open-weight models.} Best/mean/range are computed over models within each MSC. Range is best$-$worst in percentage points.}
  \label{tab:msc-trends-filled}
  \setlength{\tabcolsep}{4pt}
  \begin{tabular}{@{}l l r l
                  S[table-format=2.1]
                  S[table-format=2.1]
                  S[table-format=2.1]@{}}
    \toprule
    Category & MSC (area) & {$n$} & Winner &
    \multicolumn{1}{c}{Best (\%)} &
    \multicolumn{1}{c}{Mean (\%)} &
    \multicolumn{1}{c}{Range (pp)} \\
    \midrule
    \textit{Uniformly challenging} & 16 (Rings/algebras)  &  24 & \texttt{Gemini-3-Pro}              & 48.6 & 14.6 & 47.2 \\
                                 & 52 (Convex/discrete) &  30 & \texttt{Gemini-3-Pro}              & 57.8 & 24.7 & 57.8 \\
    \addlinespace
    \textit{Uniformly easier}     & 40 (Series/summ.)    &  36 & \texttt{Grok-4.1-Fast}             & 80.6 & 69.8 & 52.8 \\
                                 & 11 (Number theory)   & 269 & \texttt{Gemini-3-Pro}              & 71.9 & 57.8 & 40.6 \\
                                 & 26 (Real functions)  & 115 & \texttt{Gemini-3-Pro}              & 75.1 & 59.4 & 49.6 \\
    \addlinespace
    \textit{High-disagreement}    & 60 (Probability)     &  24 & \texttt{Grok-4.1-Fast}             & 59.7 & 45.7 & 55.6 \\
                                 & 51 (Geometry)        &  95 & \texttt{Grok-4.1-Fast}             & 65.3 & 48.0 & 46.7 \\
                                 & 5 (Comb.)            & 131 & \texttt{Gemini-3-Pro}              & 51.9 & 31.2 & 48.6 \\
                                 & 15 (Linear algebra)  &  33 & \texttt{GPT-OSS-120B} (hard, 81920) & 55.6 & 38.7 & 42.4 \\
    \bottomrule
  \end{tabular}
\end{table}


\section{Human team profiles}
\label{app:human_teams}

We recruited five teams (A--E) whose profiles are summarized in Table~\ref{tab:human_teams} and whose per-team performance is plotted in Figure~\ref{fig:ai_human_split}. The narrative descriptions below give the full credential breakdown for each team and, where available, the team's collaboration strategy and attempt-vs-solve counts.

\paragraph{IRB applicability and participant risk.}
The human baseline was designed as an objective mathematical problem-solving exercise: participants solved benchmark questions with well-defined answers, and scoring was based on mathematical correctness rather than subjective human preference, alignment judgments, personal attitudes, or sensitive information. Participation involved ordinary time burden from a compensated contest-style evaluation, with no deception, intervention, or collection of private behavioral or medical data. On this basis, we treat the activity as not subject to IRB review.

\paragraph{CS Major (IMO exp.) (A).}
This team comprises master's and undergraduate students who had extensive olympiad experience in high school but pursued computer science instead of pure mathematics at university. The team includes 7 IMO Honorable Mentions, 2 IMO Bronze medalists, 1 EGMO Bronze medalist, and 1 APMO Bronze medalist.

\paragraph{Math Major (IMO exp.) (B).}
This team consists of current mathematics undergraduates with a strong contest background including 1 WMTC Gold, 1 APMO Bronze, 1 IMO Silver, 1 KMS Silver, and 1 KMS Gold. The team emphasized division of labor combined with cross-checking, attempting 38 problems and solving 30 correctly.

\paragraph{Math Major (IMO Gold) (C).}
This team combines elite olympiad performance with substantial competitive programming experience including 2 IMO Gold, 1 IMO Silver, 1 APMO Bronze, 1 APMO Honorable Mention, 2 KMS Gold, 2 ICPC Seoul Gold, and 1 ICPC Bronze. The team emphasized parallelism and internal competition, attempting 44 problems and solving 25 correctly.

\paragraph{Math Major (D).}
This team is more informatics- and programming-oriented while retaining a contest mathematics background, including SIMC 2.0 Champion, 1 KMS Silver, 2 KOI Bronze, 1 KOI Silver, 1 ICPC Asia Pacific Bronze, and 1 ICPC Seoul Silver.

\paragraph{Math Researchers (E).}
This team consists of five PhD holders spanning mathematics and computer science, with most having completed their undergraduate studies in mathematics and having published mathematical research.

\end{document}